\documentclass{article}

\usepackage[final]{corl_2018} % Uncomment for the camera-ready ``final'' version

\usepackage{amsmath}
\usepackage{graphicx}

\usepackage{bm}
\usepackage{varwidth}
\usepackage{algorithm}
\usepackage[noend]{algpseudocode}

\usepackage{wrapfig}
\usepackage{capt-of}% or \usepackage{caption}
\usepackage{booktabs}

\title{Few-Shot Goal Inference for \\ Visuomotor Learning and Planning}

\author{
  Annie Xie, Avi Singh, Sergey Levine, Chelsea Finn\\
  University of California, Berkeley 
}

\begin{document}
\maketitle

%===============================================================================

\begin{abstract}
Reinforcement learning and planning methods require an objective or reward function
that encodes the desired behavior. Yet, in practice, there is a wide range of scenarios where an objective is difficult to provide programmatically, such as   tasks with visual observations
involving unknown object positions or deformable objects.
In these cases, prior methods use engineered problem-specific solutions, e.g., by instrumenting the environment with additional sensors to measure a proxy for the objective. Such solutions require a significant engineering effort on a per-task basis, and make it impractical for robots to continuously learn complex skills outside of laboratory settings. We aim to find a more general and scalable solution for specifying goals for robot learning in unconstrained environments.
To that end, we formulate the few-shot objective learning problem, where the goal is to learn a task objective from only a few example images of successful end states for that task.
We propose a simple solution to this problem: meta-learn a classifier that can recognize new goals from a few examples. We show how this approach can be used with both model-free reinforcement learning and visual model-based planning and show results in three domains: rope manipulation from images in simulation, visual navigation in a simulated 3D environment, and object arrangement into user-specified configurations on a real robot.
\vspace{-0.2cm}
\end{abstract}


\vspace{-0.2cm}
\section{Introduction}
\label{sec:intro}
\vspace{-0.2cm}

Reinforcement learning and planning methods assume some form of objective or reward function
that encodes the desired outcome or behavior.
There is a range of tasks where such an objective is challenging for humans to convey to robots, such as vision-based control with unknown object positions and manipulation of deformable objects, among a number of others. Even for relatively simple skills such as pouring or opening a door, prior works have hand-designed mechanisms to measure a proxy for the objective. This means that even task-agnostic methods such as reinforcement learning still require task-specific engineering to learn a task.
In contrast to robots, humans can very quickly infer and understand task goals. This ability to mentally represent an objective and what it means to accomplish a task is a critical aspect of autonomously learning complex skills, as it is the driver of learning progress. If we aim to build robots that can autonomously learn new skills in real-world environments, where external feedback comes rarely, then we must develop robots that can build an internal understanding of its goals and a general mechanism for humans to convey these goals.

One simple approach for specifying tasks is to provide an image of the goal~\cite{jagersand1995visual,deguchi1999image,watter2015embed,finn2016deep,zhu2017target,upn}, or more generally, provide an observation of one instance of success. There are a number of challenges with this approach, such as measuring the distance between the current and goal observation; but perhaps most saliently, we would like to not only encode a single instance of success, but reason about the entire space of successful behavior and generalize the high-level goal to new scenarios.
To encode such goals, we can learn a reward function~\cite{abbeel2004apprenticeship,ziebart2008maximum,finn2016guided,sermanet2016unsupervised,tung2018reward} or success classifier~\cite{pinto2016supersizing,ho2016generative,levine2016learning} that operates on the robot's observations. Rewards and classifiers can be learned respectively through inverse reinforcement learning from demonstrations and supervised learning from positive and negative examples. Yet, these methods do not solve the entire problem, as they require a considerable amount of data for acquiring an objective for a single task.

If we would like robots to autonomously learn to perform a wide range of tasks, then it is impractical to provide many examples for training a classifier or reward for each and every task. Humans can grasp the goal of a task from just one or two examples of success. Can we allow robots to do the same?
If we reuse data from a range of other tasks that we might want a robot to learn, then learning goal metrics for new tasks can be done much more efficiently. In this paper, we consider the problem of learning goal classifiers for new tasks with only a few examples by reusing data from other tasks. 

\begin{figure}
    \centering
    \vspace{-0.3cm}
    \includegraphics[width=0.8\textwidth]{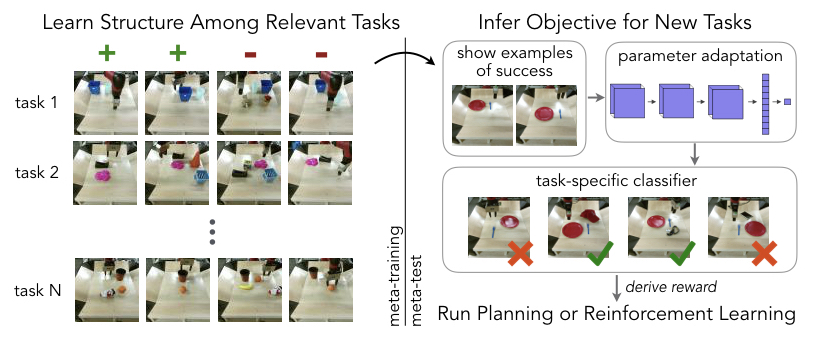}
    \vspace{-0.3cm}
    \caption{\small We propose a framework for quickly specifying visual goals. Our goal classifier is meta-trained with positive and negative examples for diverse tasks (left), which allows it to meta-learn that some factors matter for goals (e.g., relative positions of objects), while some do not. At meta-test time, this classifier can learn goals for new tasks from a couple of examples of success (right - the goal is to place the fork to the right of the plate). The reward can be derived from the learned goal classifier for use with planning or reinforcement learning.}
    \label{fig:teaser}
    \vspace{-0.3cm}
\end{figure}

Our main contribution is a framework for specifying goals to robots with minimal human supervision for each individual task. Our approach, illustrated in Figure~\ref{fig:teaser}, leverages recent work in meta-learning to enable few-shot acquisition of goal classifiers, and can provide an objective for various control optimizations, including visual model-based planning and model-free reinforcement learning. In our evaluation, we find that our approach provides a means to quickly learn objectives for simulated visual navigation, manipulation of deformable objects in simulation, and multi-stage vision-based manipulation tasks on a real Sawyer robot.

\vspace{-0.1cm}
\section{Related Work}
\label{related}
\vspace{-0.2cm}

Specifying goals is a challenge for many real-world robotics and reinforcement learning tasks. Robotic learning methods often sidestep this issue and instead hand-engineer task objectives~\cite{levine2016end,progressive_simtoreal}, often using manually instrumented environments (e.g., a thermal camera or scale to evaluate pouring~\cite{schenck2017visual,schenck2016guided,yamaguchi2015pouring},  mocap to track a pancake~\cite{pancake}, and an accelerometer on a door handle~\cite{yahya2017collective}).
We propose a more general and scalable framework for specifying a goals that does not require manual instrumentation or shaping, and instead learns to represent goals using the robot's sensors that are used to complete the task.

A number of works have proposed to specify vision-based tasks using an image or visual representation of the goal~\cite{jagersand1995visual,deguchi1999image,watter2015embed,finn2016deep,edwards2016perceptual,zhu2017target,edwards2017cross,upn}. Unlike these works, we aim to acquire a classifier that can effectively recognize successful
task executions that may not directly match an image of the goal. This enables us to recognize goals that are more abstract than an entire goal visual scene, such as relative positions of objects, approximate shapes of deformable objects, or disjunctions. 
Other works have sought to learn objectives by training a classifier~\cite{pinto2016supersizing,ho2016generative,levine2016learning} or reward function~\cite{sermanet2016unsupervised,christiano2017deep,tung2018reward}, including in the framework of inverse RL~\cite{ng2000algorithms,abbeel2004apprenticeship,ziebart2008maximum,boularias2011relative,kalakrishnan2013learning,finn2016guided,wulfmeier2016watch,firstperson_forecasting,vice}.
However, training a classifier from scratch per task has a number of challenges: modern vision methods require large training sets, and generally require both positive and negative examples. Providing many examples is onerous, and requiring users to provide negative examples is time-consuming and counter-intuitive to the average user. We aim to address both of these issues by considering the fact that we ultimately care about learning objectives of many different skills -- we can use meta-learning to share data across tasks, such that only a modest amount of data is needed for any individual task, and learn how to learn a task objective from only a handful of positive examples.

Our work builds upon the ideas of learning-to-learn~\cite{thrun,schmidhuber1987,hochreiter} and few-shot learning~\cite{lake2015human,mann,matchingnets,hugo,finn2017model,metanetworks}. 
Our general objective of quickly conveying goals to robots is related to that of prior works on one-shot imitation learning~\cite{duan_one_shot_imitation,mil,yu2018one} and recent work on meta-inverse reinforcement learning~\cite{xu2018learning,gleave2018multi}, both of which strive to learn behavior from one or a few demonstrations either through direct imitation or by first learning a reward.
Unlike these methods, our aim is to acquire goal definitions from example observations without access to a full demonstration. That is, our method only requires examples of \emph{what} successful completion of the task looks like, rather than \emph{how} to do it.
In addition to these imitation and inverse RL methods, meta-learning has been used for control in the context of reinforcement learning~\cite{duan2016rl,wang2016learning,finn2017model}. Our work focuses on fast objective learning, which could be used in conjunction with meta-reinforcement learning for acquiring behavior quickly from a few examples of success.

\vspace{-0.1cm}
\section{Generalized Framework for Few-Shot Goal Inference}
\label{sec:method}
\vspace{-0.2cm}

\newcommand{\task}{\mathcal{T}}
\newcommand{\data}{\mathcal{D}}
\newcommand{\obs}{\mathbf{o}}
\newcommand{\out}{y}
\newcommand{\posdata}{\data^+}
\newcommand{\testdata}{\data^\text{test}}
\newcommand{\loss}{\mathcal{L}}
\newcommand{\learner}{f_\text{L}}

We aim to develop a framework that makes it easy to specify objectives for new tasks in a way that (1) reduces the manual engineering efforts for specifying a new task to a robot, making it fast and easy to convey goals to robots and (2) is applicable to a wide range of problems and robot learning methods. Since we will be using camera images, we will focus on problems with visually discernible goals, such as object manipulation and visual navigation tasks; but in principle, sensors other than cameras could be used.

Assuming we ultimately want to convey many different tasks to a robot, we consider a multi-task problem setting where we have a modest number of success/failure examples for a large number of tasks. We will use this data for meta-training a classifier such that, at test time, we can learn a goal classifier for a new task from only a few examples of success. 
By doing this, we minimize the amount of data needed for any particular task and make it possible to easily and quickly convey the goal of any new task. Following the meta-learning literature~\citep{finn2018thesis}, we will consider the most general notion of a task, which can encapsulate a different objective, a different domain, a different environment, or a combination thereof.
For full generality and minimal task-specific or domain-specific engineering, we need to be able to evaluate the objective using the same observation that the robot uses to solve the task, rather than external sensors or privileged information that is not available during deployment. To satisfy this requirement, we will learn success classifiers that operate directly on the robot's observation space.
With these two design decisions in mind, we will next formalize the general learning problem and discuss our high-level solution that uses meta-learning.

\vspace{-0.1cm}
\subsection{Problem Set-up}
\vspace{-0.1cm}

% Problem set-up
Formally, we consider a goal classifier $\hat{\out} = f(\obs)$, where $\obs$ denotes the robot's observation, such as a camera image, and $\hat{\out} \in [0,1]$ indicates the predicted probability of the observation being of a successful outcome of the task. Such a classifier can be used for specifying the goal to reinforcement learning or planning algorithms, by deriving a reward function from the classifier's predictions; we discuss this in more detail in Section~\ref{sec:derive_reward}.
Our aim is to learn a goal classifier from a few positive examples of success for a new task $\task_j$, as positive examples of success are intuitive for a human to provide and, in a sense, are the minimal piece of information needed to convey a task goal.
Hence, we will be given a dataset $\posdata_j$ of $K$
positive examples of success for a task $\task_j$: $\data_j:=\{(\obs_k, 1) | k = 1...K\}_j$, and our goal is to infer a classifier for the conveyed task.
How might we go about learning to infer goal classifiers for new tasks from only $K$ positive examples? To do this, we will explicitly train a model for the ability to infer goal classifiers for a wide range of previous tasks, $\{ \task_i \}$.
In particular, we assume a small dataset $\data_i$ for each task $\task_i$, where each dataset consists of both examples of success and not success: $\data_i:= \{(\obs_n,\out_n) | n=1...N \}_i$. 

One natural question at this point is: what can the model learn from the meta-training set that allows it to infer goals for new tasks more effectively than learning each task from scratch? If each task involves a completely distinct visual concept, without any shared structure across tasks, then it seems unlikely that the model will acquire useful knowledge from meta-training. However, practical real-world goals often share many patterns: object rearrangement tasks depend strongly on relative positioning of objects to each other, and are agnostic to the pose of the robot. Tasks that involve placing objects in containers depend on whether or not an object is inside the container, but not the container's position. By extracting such patterns from meta-training tasks, our method should be able to acquire structurally related meta-test tasks efficiently.

\vspace{-0.1cm}
\subsection{Meta-learning for Few-Shot Goal Inference}
\vspace{-0.1cm}

% General solution.
To solve the above learning problem, we propose to learn how to learn classifiers for a task from a few positive examples of that task, by using full supervision at the meta-level from both positive and negative examples. Then, at test time, we can effectively learn a classifier from only positive examples. 
Across all of the tasks in the set of training tasks $\{\task_i\}$, we will train a learner $\learner$ to learn a goal classifier $g_i$ from a dataset of positive examples $\posdata_i$ and make predictions about new observations $\obs$:
$$
g_i(\obs) = \learner(\posdata_i, \obs; \theta)
$$
where we use $\posdata_i$ to denote a dataset of $K$ examples sampled uniformly from the positive examples in $\data_i$. To train the meta-learner parameters $\theta$, we will optimize the learned classifier for its ability to accurately classify new examples in $\data_i$, both positive and negative.
In particular, we will optimize the following objective:
\begin{align}
\min_\theta ~\sum_i \sum_{(\obs_n, y_n) \in \testdata_i} \loss(\out_n, g_i(\obs_n)) = \min_\theta ~\sum_i
\sum_{(\obs_n, \out_n) \in \testdata_i} 
\loss(\out_n, \learner(\posdata_i, \obs_n; \theta)) 
\label{eq:metaobj}
\end{align}
where $\posdata_i$ is defined as above, $\testdata_i$ is sampled uniformly from $\data_i \backslash \posdata_i$, and the loss function $\loss$ is the standard cross entropy loss that compares the classifier's predictions to the labels. The datapoints $\testdata$ are distinct from $\posdata$ so that we train for good generalization.
At test time, after learning $\learner$, we are presented with  examples of success  $\posdata_j$ for a new task $\task_j$. We can use this data infer a task-specific goal classifier $g_j(\cdot) = \learner(\posdata_j, \cdot; \theta)$.
This classifier provides an objective, or part of an objective (as we discuss in the next section), for reinforcement learning or planning.  
We refer to this approach as few-shot learning of objectives (FLO). The algorithms underlying the meta-training optimization and test-time procedure are outlined in Algorithms~\ref{alg:train} and~\ref{alg:test}.

\begin{figure}[ttt!]
\begin{minipage}[t]{0.62\textwidth}
\begin{algorithm}[H]
    \caption{Few-Shot Learning of Objectives (FLO)}
    \small
    \label{alg:train}
    \begin{algorithmic}[1]
    \Require for each task $\task_i$, a dataset of example successes and failures: $\data_i := \{(\obs_n, \out_n)  \}_i   ~~\forall~ i$ 
    \State randomly initialize learner parameters $\theta$
    \While{not done}
    \State Sample training task $\task_i$ (or minibatch)
    \State Sample examples of success $\posdata_i \sim \data_i$
    \State Sample test examples $\testdata_i \sim \data_i \backslash \posdata_i $
    \State \begin{varwidth}[t]{\linewidth} Learn goal classifier from positive examples $g_i(\cdot) = \learner(\posdata_i, \cdot; \theta)$ \end{varwidth}
     \State \begin{varwidth}[t]{\linewidth} Update learner parameters $\theta$ according to Eq.~\ref{eq:metaobj} using $\testdata_i$ \end{varwidth}
    \EndWhile
    \end{algorithmic}
\end{algorithm}
\end{minipage}
\begin{minipage}[t]{0.37\textwidth}
\begin{algorithm}[H]
    \caption{FLO test-time}
    \small
    \label{alg:test}
    \begin{algorithmic}[1]
    \Require examples of success $\posdata_j$ for new task $\task_j$
    \Require learned $\theta$
    \State \begin{varwidth}[t]{\linewidth} Infer goal classifier: \\ $g_j(\cdot) = \learner(\posdata_j, \cdot; \theta)$ \end{varwidth}
     \State \begin{varwidth}[t]{\linewidth}  Run RL or planning, using reward/cost derived from $g_j$  \end{varwidth}
    \end{algorithmic}
\end{algorithm}
\end{minipage}
\vspace{-0.3cm}
\end{figure}

As discussed by~\citet{universality}, the view of meta-learning as learning the mapping $\learner(\data, \obs; \theta) \rightarrow \hat{\out}$ is general to a number of different meta-learning algorithms, including recurrent models~\cite{mann}, learned optimizers~\cite{hugo}, and gradient-based methods~\cite{finn2017model}.
Hence, this framework can be combined with any of such meta-learning algorithms for few-shot classifier learning.

\vspace{-0.1cm}
\section{Few-Shot Goal Inference for Learning and Planning}
\label{sec:method}
\vspace{-0.2cm}
\newcommand{\caml}{f_\text{CAML}}

Having presented the general framework of few-shot goal inference, we will discuss our particular meta-learning implementation, mechanisms needed to mitigate exploitation of the learned objective by the controller, and mechanisms for specifying compound tasks by joining classifiers.

\subsection{Concept Acquisition for Goal Classifiers}

While the above framework is general to any meta-learning approach, we would use a method that can efficiently learn to learn, to avoid collecting very large amounts data for meta-training. As a result, we choose to build upon model-agnostic meta-learning (MAML)~\cite{finn2017model}, which incorporates the structure of gradient descent for efficient meta-learning. In particular, MAML learns an initial parameter setting $\theta$ of the model $f_\text{MAML}$ such that one or a few steps of gradient descent with a few examples leads to parameters $\theta'$ that generalize well to new examples from that task. \citet{caml} extended MAML for learning new concepts from only positive examples, referred to as concept acquisition through meta-learning (CAML), akin to how humans learn new concepts. We directly apply CAML to the setting of acquiring binary success classifiers from positive examples. At test time, the learner uses gradient descent to adapt the meta-learned parameters $\theta$ to a dataset of positive examples $\posdata_j$ for task $\task_j$:
$$
g_j(\obs) 
= \learner(\posdata_j, \obs; \theta) 
=\caml\big(\obs; \theta_j' \big)
=\caml\big(\obs; \theta-\alpha \nabla_\theta \!\!\! \sum_{(\obs_n, \out_n)\in \posdata_j} \loss (\out_n, \caml(\obs_n; \theta)\big)
$$
where $\loss$ is the cross-entropy loss function, $\alpha$ is the step size, and $\theta_j'$ denotes the parameters updated through gradient descent on task $\task_j$. We only write out one gradient descent step for convenience of notation; in practice, more may be used. Then, meta-training takes into account this gradient descent adaptation procedure, training for the initial parameters as follows:
$$
\min_\theta \sum_i \sum_{(\obs_n, y_n) \in \testdata_i} \loss(\out_n, \caml(\obs_n; \theta_i')) 
$$
We optimize this objective using Adam~\cite{adam} on the initial parameters $\theta$.
In our experiments, we consider vision-based tasks where $\obs$ is an RGB image. We use a learner $\caml$ that is represented by a convolutional neural network with RGB image inputs. We provide details on the architecture and other implementation details in Section~\ref{sec:result}.

\subsection{Deriving Rewards from Classifier Predictions}
\label{sec:derive_reward}

After meta-learning, a goal classifier can be inferred for a new task $\task_j$ by running one gradient descent step from the initial parameters $\theta$ with respect to a few examples of success, $\testdata_j$. The inferred goal classifier predicts the probability of an observation $\obs$ being successful at performing task $\task_j$. To use this prediction as an objective for planning or reinforcement learning, we need to convert it into a reward function. One simple approach would be to treat the probability of success as the reward for that observation. We find that this works reasonably well, but that the predictions produced by the neural network are not always well-calibrated.  
To reduce the effect of false positives and mis-calibrated predictions, we use the classifier conservatively by thresholding the predictions so that reward is only given for confident successes. Below this threshold, we give a reward of 0 and above this threshold, we provide the predicted probability as the reward. The threshold is chosen via cross-validation on a set of validation tasks.

\subsection{Cascading Classifiers for Compound Tasks}
\label{sec:cascade}

To provide objectives for more complex tasks, we can join multiple classifiers. For example, if we train a classifier to recognize if a particular relative position of two objects is achieved, then we can use multiple classifiers to achieve a particular configuration of multiple objects, like a table setting. To achieve this, we provide a few positive examples $\posdata_j$ of each task $\task_j$ that we would like to achieve, and then infer the classifier for each task. To perform the sequence of tasks, we cascade the inferred classifiers -- iteratively setting each objective, and running the planner or policy for each one until the classifier indicates that the subtask has been completed. In our experiments, we illustrate how this cascading technique can be used to maneuver three objects into a desired configuration.

\vspace{-0.1cm}
\section{Experiments}
\label{sec:result}
\vspace{-0.2cm}

To study the generality of our approach, we evaluate our learned objectives with both vision-based planning and reinforcement learning methods, with an emphasis on tasks that have visually nuanced goals that are difficult to specify manually. Our experiments focus on object arrangement, rope manipulation, and visual navigation tasks, but our approach is suitable for any domain where a few successful observations can sufficiently communicate the task. Code for training the few-shot success classifier and videos of our results are available on the supplementary website\footnote{The supplementary website is at \url{https://sites.google.com/view/few-shot-goals}}.

Since our goal is to provide an easy way to compute an objective for new tasks from a few observations of success for that task, we compare our approach to a few alternative and prior methods for doing so under the same assumptions as our method:

\textbf{Pixel distance}: Given one observation of success, a naive metric is to measure the $\ell_2$ distance between the current observation and the successful observation. Will this can sometimes perform well with low-dimensional observations, this approach cannot capture invariances that a goal classifier can represent, nor does it scale well to visual or other high-dimensional sensory observation spaces.

\textbf{Latent space distance}: An alternative approach is to measure the distance between current and goal observations in a learned latent space~\cite{watter2015embed,finn2016deep}. For learning the latent space, we train an autoencoder~\cite{lange2012autonomous,finn2016deep} on the meta-training data used for our approach. We expect this metric to be better grounded than distances in pixel space, but still cannot capture invariances across and within goals.

\textbf{Oracle}: In our rope domain, we derive an objective using the ground truth state of the environment, which is generally not available in the real world. This provides an upper bound on performance.

For all tasks in our evaluation, we consider the 5-shot learning setting, where 5 examples of success are provided to method for conveying the goal of the test task.
\begin{wrapfigure}{R}{0.6\textwidth}
    \vspace{-0.6cm}
    \includegraphics[width=0.6\textwidth]{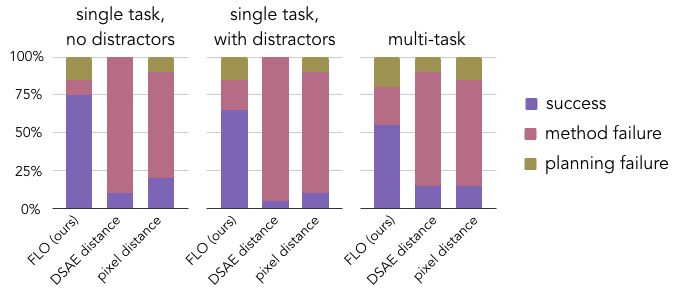}
    \includegraphics[width=0.6\textwidth]{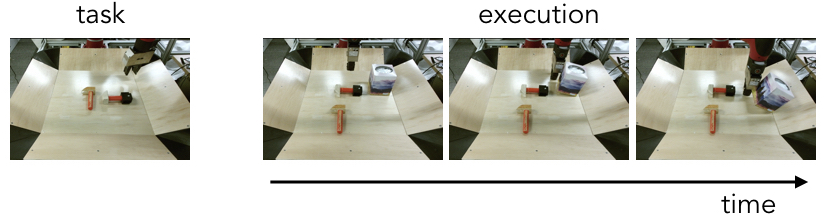}
    \vspace{-0.77cm}
    \caption{\small Top: quantitative performance of visual planning across different goal specification methods: ours, DSAE~\cite{finn2016deep}, and pixel error. Where possible, we include break down the cause of failures into errors caused by inaccurate prediction or planning and those caused by an inaccurate goal classifier. Bottom: visualization of the most common failure  case --- the presence of novel distractor objects. More data or data augmentation with a wider range of distractors could help mitigate this source of failure.}
    \label{fig:planning_results}
    \vspace{-0.5cm}
\end{wrapfigure}
For both the pixel and latent space distance metrics above, we take the minimum distance across these five examples, allowing the optimization to try to match the closest goal observation. Full experimental details and neural network architecture information can be found in Appendices~\ref{app:arch} and~\ref{app:data}.

\begin{figure}[t]
    \centering
    \vspace{-0.4cm}
    \includegraphics[width=\textwidth]{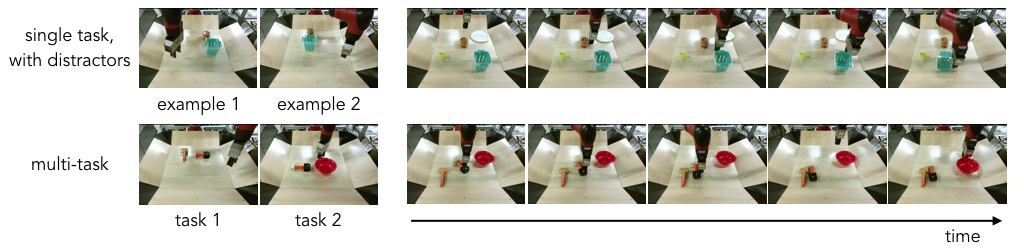}
    \vspace{-0.75cm}
    \caption{\small Object arrangement performance of our method with distractor objects and with two tasks. The left shows a subset of the 5 positive examples that are provided for inferring the goal classifier(s), while the right shows the robot executing the specified task(s) via visual planning.}
    \vspace{-0.4cm}
    \label{fig:qual_plan}
\end{figure}

\vspace{-0.1cm}
\subsection{Visual Planning for Object Arrangement}
\vspace{-0.1cm}

We study a visual object arrangement task, where different goals correspond to different relative arrangements of a pair of objects. For this setting, we use a Sawyer robot and use the planning method developed by~\citet{sna} to optimize actions with respect to the learned objective. This planning approach learns a video prediction model from self-supervised data, and uses a sampling-based optimization with iterative replanning (MPC) to select sequences of actions at each timestep that lead to desirable futures. We evaluate our learned classifier on the predictions made by the video prediction model and derive the cost used for planning using the approach described in Section~\ref{sec:derive_reward}.

We evaluate all approaches in three different experimental settings. In the first setting, the goal is to arrange two objects into a specified relative arrangement. The second setting is the same, but with distractor objects present. In the final, most challenging setting, the goal is to achieve two tasks in sequence. As described in Section~\ref{sec:cascade}, we provide positive examples for both tasks, infer the classifier for both task, perform MPC for the first task until completion, followed by MPC for the second task. 
To evaluate the ability to generalize to new goals and settings, we use novel, held-out objects for all of the task and distractor objects in our evaluation.

\begin{figure}
    \vspace{-0.2cm}
    \hspace*{-0.25cm}\includegraphics[width=\textwidth]{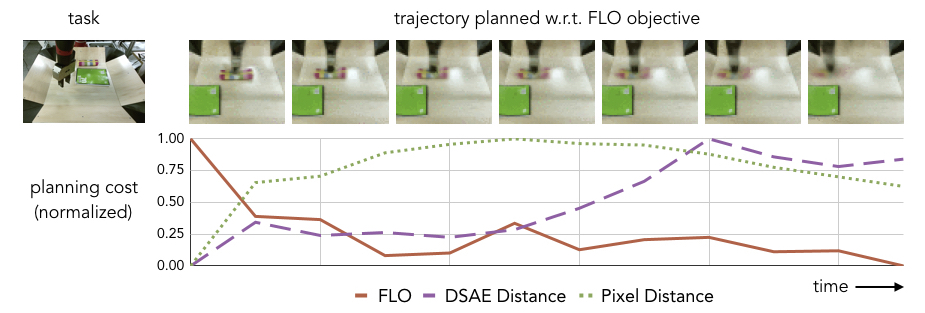}
    \vspace{-0.3cm}
    \caption{\small Comparison of the normalized cost determined by each method (bottom). We evaluate each metric on the video prediction corresponding to the trajectory planned with the FLO objective (top).}
    \label{fig:cost_comparison}
    \vspace{-0.2cm}
\end{figure}

We qualitatively visualize the evaluation in Figure~\ref{fig:qual_plan}. On the left, we show a subset of the five images provided to illustrate the task(s), and on the left, we show the motions performed by the robot. We see that the robot is able to execute motions which lead to a correct relative positioning of the objects.
We quantitatively evaluate each method across 20 tasks, including $10$ unique object pairs. The results, shown in Figure~\ref{fig:planning_results}, indicate that prior methods for learning distance metrics struggle to infer the goal of the task, while our approach leads to substantially more successful behavior on average. Aiming to understand the results further and the fidelity of the inferred goal classifiers, we measure whether the failures are caused by the planner or the classifier. In particular, if the classifier predicted that the final image was a success that was not, or if the classifier classifies a correct plan as not successful, then the fault is on the classifier. Otherwise, the fault is the planner, e.g. if the planner finds an action plan that looks successful according to the video prediction model, but in reality is not. From this analysis, we observe that about half of the single task failures are caused by the planner / predictive model. This suggests at least 85\% of the tasks are achievable by the planner with a suitable metric. We further analyze a common failure mode in Figure~\ref{fig:planning_results} and provide a direct comparison of each method's cost function in Figure~\ref{fig:cost_comparison} and Appendix~\ref{app:analysis}.

\vspace{-0.1cm}
\subsection{Rope Manipulation with Reinforcement Learning}
\vspace{-0.1cm}

We next study a setting where the goal is to learn to manipulate a rope of pearls into a particular shape, as specified by a few images. Here, we aim to evaluate how our objective and others can be used with the cross-entropy method~\cite{rubinstein2013cross}. For this setting, we use a parallel-jaw gripper as the manipulator, and our experiments are carried out in the Mujoco simulator~\cite{todorov2012mujoco}.

\begin{figure}[t]
    \centering
    \includegraphics[width=\textwidth]{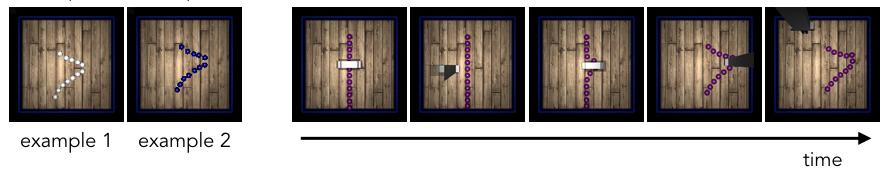}
    \vspace{-0.8cm}
    \caption{\small Rope manipulation policy roll-outs. Left:  a subset of the 5 positive examples provided for inferring the goal classifier. Right: the robot executing the policy learned via RL with the learned objective.}
    \vspace{-0.5cm}
    \label{fig:qual_rope}
\end{figure}

We use 30 tasks for testing, with five success images used per task for inferring the goal. Example tasks are illustrated in Figure~\ref{fig:rope_results}. For evaluation and the oracle, we need to devise a distance metric. We do so by translating the current and goal rope to have the same center of mass, and then measuring the average distance between corresponding pearls. This metric is used only by the oracle (as reward), and to provide a fast, automatic evaluation. In our results, shown in Figure~\ref{fig:rope_results}, we first notice that pixel distance provides a reasonably good metric for the tasks. This is not all that surprising, since the simulated images are clean, with a fixed background. As shown in the previous experiments, pixel distance does not provide a good metric when using real images with novel objects, so we do not expect these results to transfer to real world settings. We also observe that FLO performs substantially better than using  a distance metric derived from the latent space of an autoencoder, but does not reach the performance of the oracle distance metric. In most tasks where the policy performed poorly using FLO, we observed that RL was able to successfully optimize the objective, indicating that it was exploiting inaccuracies in the classifier. Integrating frameworks for mining negatives and iteratively retraining the objective (such as those presented in Fu et al.~\cite{vice}) is an interesting direction for future work. We show a qualitative example in Figure~\ref{fig:qual_rope}.

\begin{table}[ht]
\vspace{-0.1cm}
\begin{minipage}[b]{0.63\textwidth}
    \centering
    \includegraphics[width=0.95\textwidth]{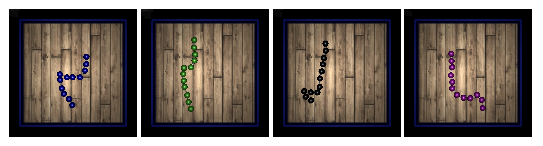}
    \label{fig:rope_tasks}
\end{minipage}
\hfill
\begin{minipage}[b]{0.36\textwidth}
\begin{center}
{\footnotesize
\begin{tabular}{lcc}
\toprule
\textbf{method} & \multicolumn{2}{c}{\textbf{median distance}} \\
\midrule
pixel distance  & 0.54 (0.34, 0.70) \\
AE distance & 0.59 (0.50, 0.69) \\ 
FLO (ours) & 0.27 (0.18, 0.34) \\
\midrule
oracle & 0.16 (0.13, 0.23) \\
\bottomrule
\end{tabular}
}
\end{center}
\vspace{-0.1cm}
\label{tbl:rope}
\end{minipage}
\vspace{-0.4cm}
\captionof{figure}{\small Left: Illustration of four different test tasks. Right: Median distance across rope manipulation tasks (lower is better),  comparing each method for deriving a reward from images. We compare RL from full state and from visual features. 
Numbers in parentheses indicate the $25^\text{th}$ and $75^\text{th}$ percentiles.}
\label{fig:rope_results}
\end{table}

\vspace{-0.1cm}
\subsection{Visual Navigation with Reinforcement Learning}
\vspace{-0.1cm}

We lastly consider a visual navigation task in VizDoom, a first-person 3D navigation environment. Here, the goal is to learn to navigate to a particular object that is specified by the few observations of success. We use DQN~\cite{Mnih15} to optimize a policy with respect to the learned reward functions. The observation space for the robot consists of its spatial location and orientation. There are four discrete actions: move forward, turn left, turn right and no action. Each episode is 30 timesteps long, and we run the algorithm for 50K timesteps. Since we found it difficult to engineer a success metric for this task, evaluations are performed by a human. For the same reason, we do not compare to an oracle that receives the true success metric as supervision, and we only compare to using pixel and latent space distance methods described above. For evaluation, a trial is successful if the robot visits a state in which the target object is in the lower third of the agent's field of view.

\begin{figure}[t]
    \centering
    \vspace{-0.4cm}
    \includegraphics[width=\textwidth]{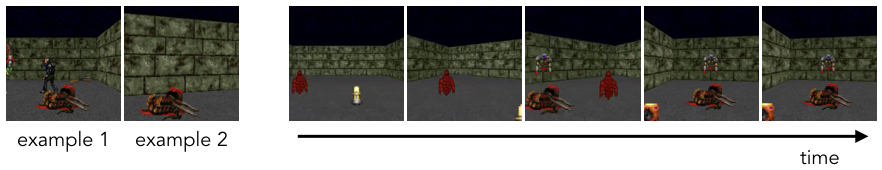}
    \vspace{-0.7cm}
    \caption{\small Visual navigation policy roll-out. Left: 2 of the 5 positive examples provided for inferring the goal classifier. Right: the agent executing the policy learned via RL using the learned objective.}
    \vspace{-0.5cm}
    \label{fig:qual_nav}
\end{figure}

We evaluate each method across 9 tasks, with 5 trials for each task for a total of 45 trials. We illustrate examples of tasks in Figure~\ref{fig:nav_results}. The reward is inferred from 5 examples of success provided for each task. Because of the limited number of objects available in VizDoom, we do not evaluate on held-out target objects, but instead evaluate on novel combinations of target and distractor objects in new configurations. As seen in Figure~\ref{fig:nav_results}, our method is able to learn a success metric and use it to navigate to an object $77.8\%$ of the time, while the alternative distance metrics have much more difficulty encoding the correct task, achieving $33.0\%$ and $44.0\%$ success respectively. We show a qualitivate example in Figure~\ref{fig:qual_nav}.

\begin{table}[ht]
\vspace{-0.1cm}
\begin{minipage}[b]{0.63\textwidth}
    \centering
    \includegraphics[width=0.92\textwidth]{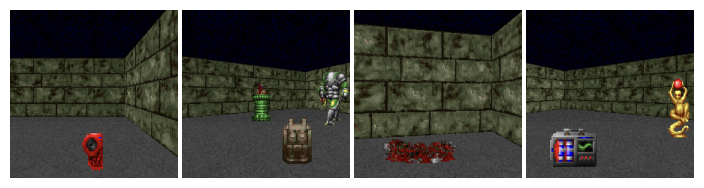}
    \label{fig:nav_tasks}
\end{minipage}
\hfill
\begin{minipage}[b]{0.36\textwidth}
\vspace{-0.1cm}
\begin{center}
{\footnotesize
\begin{tabular}{lc}
\toprule
\textbf{method} & \textbf{success rate} \\
\midrule
pixel distance & 33.0\%  \\
AE distance & 44.0\% \\
FLO (ours) & 77.8\% \\ 
\bottomrule
\end{tabular}
}
\end{center}
\label{tbl:nav}
\end{minipage}
\vspace{-0.3cm}
\captionof{figure}{\small Left: Illustration of four different test tasks. Right: Success rate of each method evaluated on the test visual navigation tasks. We compare our method to autoencoder and pixel distance metrics for deriving the reward for reinforcement learning.}
\label{fig:nav_results}
\vspace{-0.45cm}
\end{table}

\vspace{-0.1cm}
\section{Conclusion}
\label{sec:conclusion}
\vspace{-0.2cm}

In this work, we proposed a framework for learning to learn task objectives from a few examples of success, motivated by the challenge of specifying goals in real world settings. We discussed how to derive a reward function from our inferred goal classifier, and how our task goal classifiers can be combined for forming more complex, compound goals. Finally, we showed how our framework for inferring goals can be combined with both  reinforcement learning and planning for vision-based skills such as maneuvering a rope into a given shape in simulation, navigating to a specified target in a simulated 3D environment, and rearranging multiple previously-unseen objects in the real world.

%===============================================================================

% The maximum paper length is 8 pages excluding references and acknowledgements, and 10 pages including references and acknowledgements

\clearpage
% The acknowledgments are automatically included only in the final version of the paper.
\acknowledgments{ 
We thank the anonymous reviewers who gave useful comments. This work was supported by the NSF through IIS-1651843, IIS-1614653, and IIS-1700696, an ONR Young Investigator Program Award, ARL DCIST CRA W911NF-17-2-0181, and Berkeley DeepDrive. We also acknowledge generous equipment and computational resource support from NVIDIA, Google, and Amazon.
}

%===============================================================================

{
\footnotesize
\bibliography{references}  % .bib
}

\appendix
\part*{Appendix}
\section{Architecture Details}
\label{app:arch}
Our model $\caml$ is represented by a convolutional neural network with RGB image inputs. The network consists of three convolutional layers with 32 3$\times$3 filters and stride 2, each followed by layer normalization and a ReLU non-linearity. A spatial soft-argmax operation extracts spatial feature points from the final convolution layer. These features are passed through one fully-connected layers with 50 units and ReLU non-linearities, followed by a linear layer to the two-dimensional softmax output. The architecture in our simulated experiments is the same, except with 16 filters in each convolution layer, feature flattening instead of a spatial soft-argmax, and three fully-connected layers instead of one. When using real images, the first convolutional layer is initialized with weights from VGG-16. 

For the autoencoder, we use a convolutional neural network with three layers and 3x3 filters, where the layers have 64, 32, and 16 filters respectively. The stride is 2 for the first layer, and 1 for subsequent layers. We follow this up with three fully connected layers, that have 200, 100 and 50 units respectively. We train this autoencoder on the entire meta-training dataset, and our target reconstructions have dimension 32x32x3. We run reinforcement learning on top of the features from the last hidden layer of this autoencoder, and keep the autoencoder weights fixed during the policy learning process. 

\section{Experimental Details}
\label{app:data}
\subsection{Object Arrangement}
To collect data for meta-training the classifier, we randomly select a pair of objects from our set of training objects, and position them into many different relative positions, recording the image for each configuration. One task corresponds to a particular relative positioning of two objects, e.g. the first object to the left of the second, and we construct positive and negative examples for this task by labeling the aforementioned images. We randomly position the arm in each image, as it is not a determiner of task success. A good objective should ignore the position of the arm. We also include randomly-positioned distractor objects in about a third of the collected images. Some of the meta-training data is illustrated in the left of Figure~\ref{fig:teaser}. In total, we acquire data for 122 tasks, with roughly 25 positive and 45 negative examples for each task. We use more negatives than positives to account for the larger space of non-goal examples. As some images are shared across tasks, the total number of unique images in the dataset is 5490.

\subsection{Rope Manipulation}
The tasks (i.e. goal shapes of the rope in this case) are generated through the following automated procedure: the manipulator takes random actions, some of which cause the rope to move. If any single pearl gets displaced by more than 10 cm from its original position, we freeze the rope position and call it a new task. We then displace the rope all over the table, randomly select a color for the rope from a fixed set of 5 colors, and add small perturbations to it to generate multiple examples for this task. This encodes the fact that we only care about the shape of the rope, and not its color nor its position on the table, and small perturbations of the shape are acceptable. We use 360 tasks for training with 30 positives per task. Negative examples for each task are defined using positive examples of other tasks. We also collect a pool of 1000 negative examples by taking random actions and share these examples across all tasks, leading to a total of 11800 unique images being used for training.

\subsection{Visual Navigation}
We generate the data to meta-train the classifier as follows: we randomly select an object to be used as the target object, initialize the target object in a random position, and place 4 randomly chosen distractor objects in different randomized positions in the room. To collect positive examples, we place the agent between 70 and 90 units away from the object with an orientation between -30 and 30 degrees with respect to the target object and record the resulting image observation as a positive example. Negative examples for each task are collected by initializing the agent 100 to 200 units away from the object with an orientation between -30 and 30 degrees. We repeat this to generate 50 positive and 50 negative examples for each of 100 tasks. We additionally use positive examples from other tasks as negative examples for a particular task, and collect a general pool of 200 negative examples shared across all tasks. For the general pool of negatives, we gather examples by taking random actions and by collecting examples facing walls and corners. In total, there are 10200 unique images in our dataset. 

\section{Comparison of Planning Costs}
\label{app:analysis}
\begin{figure}[!ht]
    \centering
    \vspace{-0.2cm}
    \includegraphics[width=\textwidth]{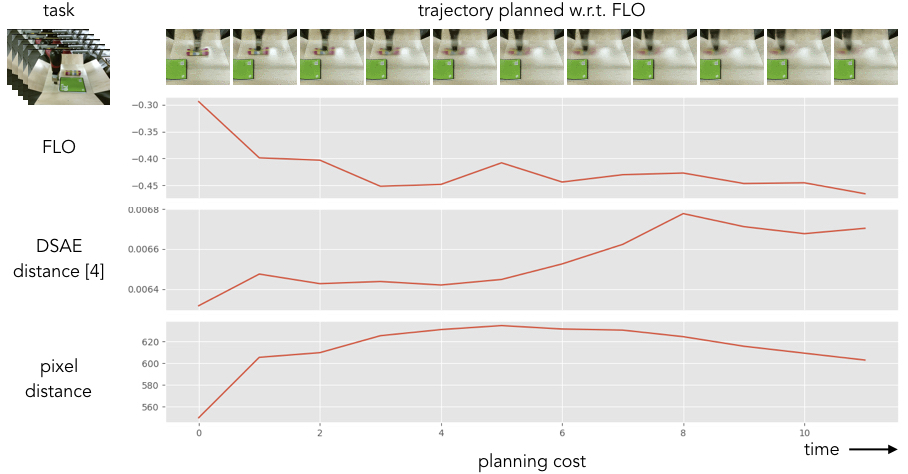}
    \vspace{-0.7cm}
    \caption{\small Trajectory proposed with FLO as the planning cost. With our method, the planner is able to discover a trajectory that enables the robot to successfully complete the specified task. For the same trajectory, the costs determined by alternative metrics do not decrease as dramatically despite each frame getting closer to success.}
    \label{fig:analysis_flo}
\end{figure}

\begin{figure}[!ht]
    \centering
    \includegraphics[width=\textwidth]{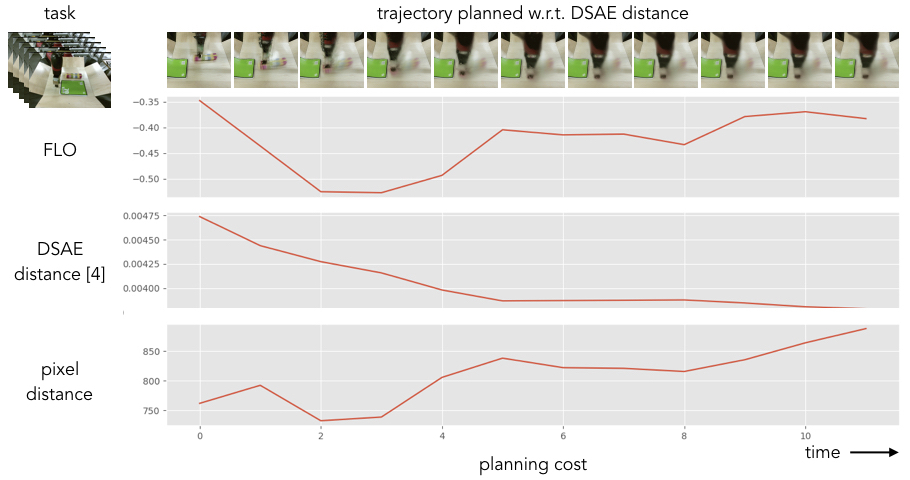}
    \vspace{-0.7cm}
    \caption{\small Trajectory proposed with DSAE distance as the planning cost. The planner using the DSAE distance tries to match the arm's position in the given example of success on the left and ignores the task entirely.}
    \label{fig:analysis_dsae}
\end{figure}

\begin{figure}[!ht]
    \centering
    \includegraphics[width=\textwidth]{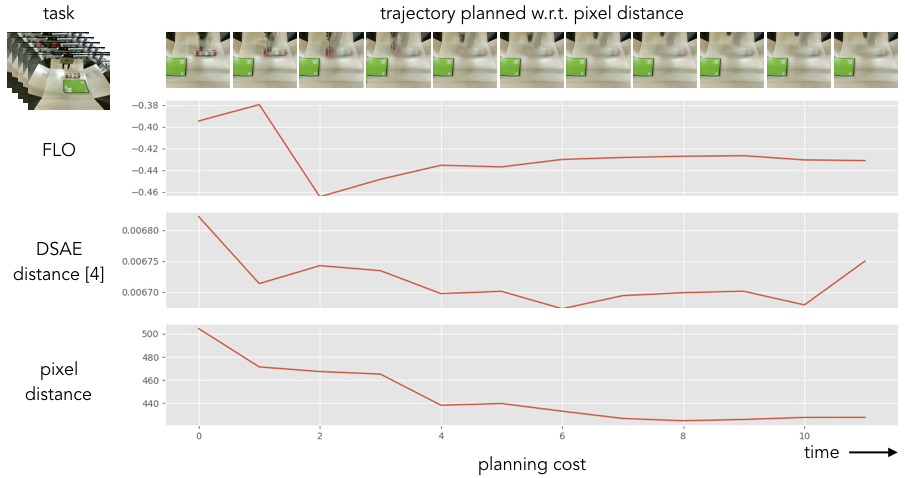}
    \vspace{-0.7cm}
    \caption{\small Trajectory proposed with pixel distance as the planning cost. The pixel distance metric also mostly focuses on the position of the arm. Instead of performing the specified task, the planner tries to move the arm to match the given example of success on the left and lower its pixel cost.}
    \label{fig:analysis_pixel}
\end{figure}

To further compare FLO to alternative success metrics, we plot the planning cost determined by each metric on trajectories planned with respect to each objective in Figures~\ref{fig:analysis_flo}, \ref{fig:analysis_dsae}, and \ref{fig:analysis_pixel}. 

\end{document}